\pdfoutput=1

\documentclass{article}
\usepackage{microtype}
\usepackage{graphicx}
\usepackage{subcaption}
\usepackage{booktabs}
\usepackage{hyperref}

\usepackage[preprint]{icml2026}

\usepackage{amsmath}
\usepackage{amssymb}
\usepackage{mathtools}
\usepackage{enumitem}
\usepackage{pifont}
\usepackage{tikz}
\definecolor{mkD}{HTML}{3B6E9C}\definecolor{mkK}{HTML}{C8502A}\definecolor{mkL}{HTML}{E39B26}\definecolor{mkF}{HTML}{A9A3C4}
\newcommand{\markD}{\tikz[baseline=-0.55ex]\draw[mkD, line width=0.6pt, fill=white] (0,0) circle (1.7pt);}
\newcommand{\markK}{\tikz[baseline=-0.55ex]\fill[mkK] (0,0) circle (1.9pt);}
\newcommand{\markL}{\tikz[baseline=-0.55ex]\fill[mkL] (0,1.9pt) -- (-2.1pt,-1.4pt) -- (2.1pt,-1.4pt) -- cycle;}
\newcommand{\markF}{\tikz[baseline=-0.55ex]\fill[mkF] (0,2.2pt) -- (1.8pt,0) -- (0,-2.2pt) -- (-1.8pt,0) -- cycle;}
\usepackage[capitalize,noabbrev]{cleveref}

\icmltitlerunning{Smallest Singular Directions Matter Disproportionately}

\begin{document}

\twocolumn[
  \icmltitle{The Smallest Singular Directions of the Unembedding Matter Disproportionately for Next-Token Prediction}

  \begin{icmlauthorlist}
    \icmlauthor{Weilun Xu}{epfl}
  \end{icmlauthorlist}

  \icmlaffiliation{epfl}{School of Computer and Communication Sciences, EPFL, Lausanne, Switzerland}
  \icmlcorrespondingauthor{Weilun Xu}{weilun.xu@epfl.ch}
  \icmlkeywords{Interpretability, Language Models, Unembedding, Fisher Information}

  \vskip 0.3in
]

\printAffiliationsAndNotice{}

\begin{abstract}
Interpreting, editing and compressing language models often involve assessing which parts of their internal representations matter for prediction. Many of these assessments rely on distance, such as reconstruction error or the length of an intervention. For next-token prediction, however, the logit differences that matter most are those among the few most plausible tokens. These differences can be small. We therefore test whether changes to the final hidden state that move the logits less also change the prediction less. Specifically, changes along the smallest singular directions of the unembedding, which maps this state to the logits, move the logits less than random changes of the same length. In open-weight language models across several families, their prediction change and their change in the next-token loss are nevertheless on average several times those of random changes. Across all singular directions, the prediction change is out of proportion to the logit change only at the smallest. These directions carry a large share of the difference between the model's leading candidate tokens. Consequently, distance-based methods undervalue these directions, and rankings by singular value place them last, although they matter disproportionately for prediction. More broadly, our study highlights that comparing internal representations by distance can miss differences that change what the model predicts.
\end{abstract}

\begin{figure}[!t]
\centering
\includegraphics[width=\linewidth]{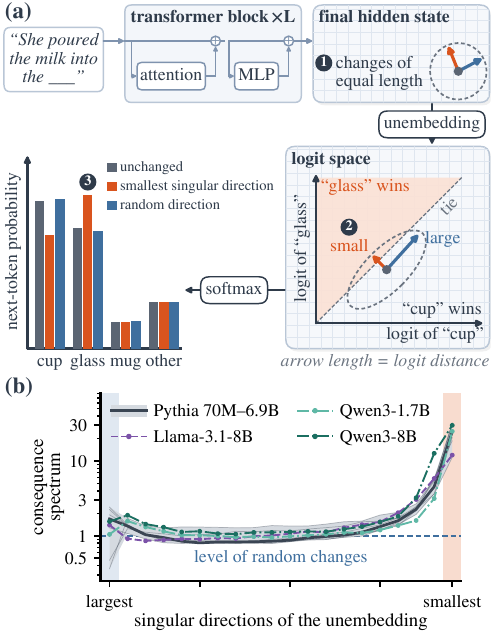}
\caption{\textbf{Changes that barely move the logits can change the prediction disproportionately.} \textbf{(a)}~\ding{182}~In an illustration, two changes of equal length are applied to the final hidden state, one along the smallest singular direction (orange) and one along a random direction (blue). \ding{183}~The unembedding maps the states to logits, shown for the two leading candidates, the most probable tokens. The orange change moves the logits less, yet it crosses the tie. \ding{184}~After the softmax, only the orange change reverses the prediction. \textbf{(b)}~The consequence spectrum (\cref{sec:instrument}): the prediction change per unit of logit distance, relative to random changes, from the largest singular value to the smallest. It stays near random over most of the spectrum and rises sharply at the smallest singular values.}
\label{fig:teaser}
\end{figure}

\section{Introduction}
\label{sec:intro}

Interpreting, editing and compressing language models often involve selecting directions in their internal representations by how much each matters for the model's predictions. Measuring this importance directly is costly, however, because every candidate change requires another evaluation of the model \citep{molchanov2019importance}. Several widely used methods therefore select by distance instead. In interpretability, the directions that the unembedding amplifies least have been described by their small effect on the logits \citep{elhage2021mathematical,stolfo2024confidence}. In editing, AlphaEdit and LEACE confine a change to a null space or minimize its size \citep{fang2025alphaedit,belrose2023leace}. In compression, low-rank methods based on the singular value decomposition keep the directions with the largest singular values, which minimizes the reconstruction error \citep{eckart1936approximation,hsu2022fwsvd}. Distance is inexpensive to compute and its optima often have closed forms. It also rests on a sound intuition: when other factors are held fixed, a small change has little effect on a network's output \citep{lecun1990optimal}.

The intuition that a small change has little effect can fail, however, when an outcome depends on a small difference between quantities that vary together. A small change to that difference can then decide the outcome, as long noted for regression on principal components \citep{jolliffe1982note}. Next-token prediction also has this structure. For example, after ``She poured the milk into the \underline{\hspace{1.2em}}'', both ``cup'' and ``glass'' may be likely. Raising both logits by the same amount leaves the choice untouched, whereas a much smaller change in their difference can reverse it. More generally, adding the same amount to every logit changes nothing, and the probability concentrates on a few plausible tokens. The prediction therefore depends mostly on the differences between their logits, and these differences can be small. We thus hypothesize that \emph{the directions along which a change moves the logits least can matter disproportionately for the prediction}. This can happen because the differences between plausible tokens may lie along these directions.

We test this hypothesis at the unembedding, the linear map from the final hidden state to the logits. It suits the test because nothing downstream can compensate for a change. In addition, its singular values measure how much it amplifies each direction of the state. We therefore group its singular directions into \emph{bands} of equal size. We then change the state within the \emph{bottom band}, which the unembedding amplifies least, and along random directions of the same length. Each change is measured in two ways. Its logit distance, how far it moves the logits, is what distance-based methods measure. Its prediction change, how much it changes the predicted distribution, is the effect on the prediction itself. The two can disagree because the logit distance counts every token equally, whereas the prediction change weights each pair of tokens by how probable both are. Comparing them from the largest singular values to the smallest gives the \textbf{consequence spectrum} (\cref{sec:instrument}): the prediction change per unit of logit distance, relative to random changes.

In ten open-weight language models from three families (Pythia, Llama and Qwen), the results support the hypothesis. Specifically, changes along the smallest singular directions move the logits less than random changes of the same length, as their small singular values imply. Nevertheless, their prediction change and their change in the next-token loss are several times those of random changes (\cref{sec:reversal}). This extends an earlier observation that removing the same band after the final layers raises the loss more than removing a random subspace of the same rank \citep{cancedda2024spectral}. The comparison also holds at other token positions, on code and in instruction-tuned models (\cref{sec:generality}). The consequence spectrum, in turn, locates the effect: it stays close to random for most directions and rises steeply at the smallest singular values (\cref{fig:teaser}). As hypothesized, these directions carry a large share of the difference between the model's leading candidate tokens, which the softmax weights most. The loss also rises with that share (\cref{sec:why}).

Distance-based methods therefore undervalue these directions, whether a model is interpreted, edited or compressed, and rankings by singular value place them last (\cref{sec:consequences}). Compression tests the cost of this ranking directly: truncating the unembedding by singular value changes the logits less than random truncation but increases the loss more. Ranking the same directions by their effect on the prediction, by contrast, increases the loss less than random truncation. Comparisons of representations by distance can likewise miss differences along these directions (\cref{sec:implications}).

Our contributions are threefold:
\begin{enumerate}[leftmargin=1.4em,topsep=2pt,itemsep=1pt,parsep=0pt,partopsep=0pt]
  \item We introduce the consequence spectrum (\cref{sec:instrument}), which compares a change's logit distance with its prediction change along a linear map's singular directions. It applies to any model with a softmax output.
  \item We find that the unembedding's smallest singular directions matter disproportionately for next-token prediction (\cref{sec:reversal}). These directions carry a large share of the difference between the model's leading candidate tokens (\cref{sec:why}).
  \item We draw the implication that a direction's importance should be measured on the prediction rather than inferred from distance. Distance-based methods undervalue these directions in interpretation, editing and compression, and rankings by singular value place them last (\cref{sec:consequences}).
\end{enumerate}

\section{The consequence spectrum}
\label{sec:instrument}

A change to a language model's final hidden state reaches the prediction through the unembedding, a linear map to the logits, and then through the softmax. We measure each such change in two ways: by its logit distance, how far it moves the logits, and by its prediction change, how much it changes the predicted distribution. These two measures, compared along the unembedding's singular directions and relative to random changes, give the consequence spectrum. For reference, \cref{tab:notation} gives an overview of the paper's notation.

\subsection{The unembedding's singular values sort the state's directions into bands}
\label{sec:unembedding}

The final hidden state $r$ is the residual stream at the last input position after the last block. This state then passes through the final normalization, which divides it by its scale and multiplies each coordinate by a learned weight $\gamma$. This normalization is LayerNorm in the Pythia models \citep{ba2016layer} and RMSNorm in the Llama and Qwen models \citep{zhang2019root}. We hold the scale at its value on the unchanged input, so that the normalized state $u$ changes linearly with $r$.

The model's unembedding matrix $W_{\mathrm{out}}$ then maps the weighted state to the logits, and we fold the weight into this matrix. In addition, we remove two components that cannot change the prediction. First, the centring $C_V = I - \mathbf 1\mathbf 1^\top/V$ over the vocabulary of $V$ tokens removes the logits' component shared by all tokens. Second, in LayerNorm models, a basis $R$ of the remaining coordinates removes the state's component along the direction that the mean subtraction removes. These steps give a single linear map, which we call the unembedding:
\begin{equation}
  W = C_V\, W_{\mathrm{out}}\, \mathrm{diag}(\gamma)\, R .
  \label{eq:readout}
\end{equation}
The logits are then $\ell = Wu$ up to biases, which a change does not alter, and a shift shared by all logits, which the softmax ignores. Each row $w_t$ of $W$ is the unembedding vector of token $t$. \Cref{app:readout} gives the normalization, the folding and the centring in full.

The singular value decomposition of $W$ orders the directions of the state by how much the unembedding amplifies them:
\begin{equation}
  W = \textstyle\sum_i \sigma_i\, a_i v_i^\top, \qquad \sigma_1 \ge \sigma_2 \ge \dots \ge 0 .
  \label{eq:svd}
\end{equation}
Here the singular directions $v_i$ are orthonormal directions of the state. In turn, the orthonormal vectors $a_i$ over the vocabulary give the pattern in which a change along each direction moves the logits. A unit change of $u$ along $v_i$ therefore moves the centred logits by $\sigma_i a_i$, a logit distance of $\sigma_i^2$. Because $W$ is centred over the vocabulary, $\sigma_i^2 = \sum_t (w_t^\top v_i)^2$ also measures how widely the tokens' unembedding vectors spread along $v_i$, with every token counted equally. We then group the directions, sorted from the largest singular value to the smallest, into twenty bands $b = 0, \dots, 19$ of equal size. Of these, the \emph{bottom band}, $b = 19$, holds the 5\% with the smallest singular values, and the \emph{top band}, $b = 0$, the 5\% with the largest.

\subsection{Logit distance and prediction change can disagree}
\label{sec:measures}

A change $\delta u$ moves the centred logits by $\delta\ell = W\,\delta u$ and changes the prediction from $p = \mathrm{softmax}(\ell)$ to $p' = \mathrm{softmax}(\ell + \delta\ell)$. Its \emph{logit distance} $D$ measures how far the logits move, and its \emph{prediction change} $K$ how much the predicted distribution changes:
\begin{equation}
  D = \|\delta\ell\|^2, \qquad K = \mathrm{KL}(p\,\|\,p').
  \label{eq:sizes}
\end{equation}

For small changes, $K$ takes a simple form that shows why the two measures differ. To second order in $\delta\ell$, it equals the \emph{Fisher term} \citep{martens2020new}, half the variance of the logit changes across tokens, with each token weighted by its probability:
\begin{equation}
  F = \tfrac12 \operatorname{Var}_{p}(\delta\ell) = \tfrac12 \textstyle\sum_t p_t\,\big(\delta\ell_t - \sum_{t'} p_{t'}\,\delta\ell_{t'}\big)^2 .
  \label{eq:fisher}
\end{equation}
In words, $F$ measures how much the logit changes differ among the tokens that hold the probability.

Written over pairs of tokens, the logit distance and the Fisher term become the same sum with different weights:
\begin{equation}
  \begin{aligned}
    D &= \tfrac{1}{V}\textstyle\sum_{t<t'} (\delta\ell_t - \delta\ell_{t'})^2,\\
    2F &= \textstyle\sum_{t<t'} p_t\, p_{t'}\, (\delta\ell_t - \delta\ell_{t'})^2 .
  \end{aligned}
  \label{eq:pairs}
\end{equation}
Each term is the squared change in the log-odds of two tokens, since $\log(p_t/p_{t'}) = \ell_t - \ell_{t'}$. In words, the logit distance counts every pair of tokens equally, and almost all pairs involve implausible tokens. The Fisher term, by contrast, weights each pair by the product of its probabilities and so counts almost only the pairs among the few plausible tokens. \Cref{app:geometry} relates both measures to established geometric constructions.

In the introduction's example, suppose the model splits its prediction evenly between ``cup'' and ``glass''. Raising one of their logits by $\epsilon$ and lowering the other by $\epsilon$ gives $D = 2\epsilon^2$ and $F = \epsilon^2/2$. The same logit distance between two improbable tokens gives $F \approx 0$, so only the prediction change tells the two changes apart.

\begin{figure*}[!t]
\centering
\includegraphics[width=\textwidth]{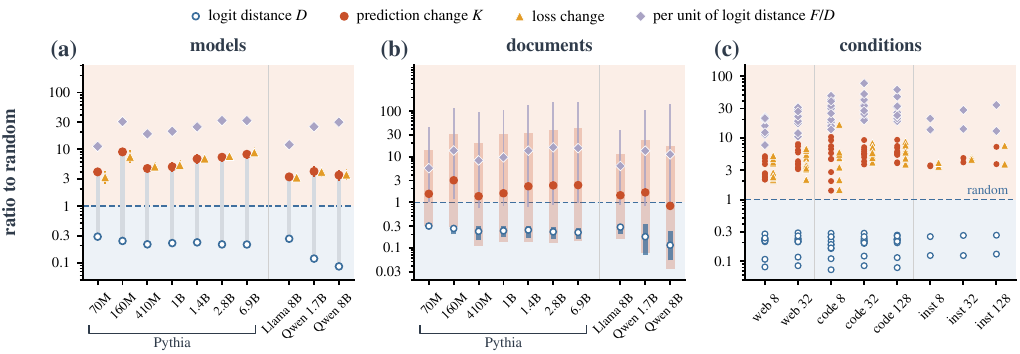}
\caption{\textbf{The bottom band changes the prediction disproportionately to its logit distance.} Each marker is the band's value divided by that of random changes of equal length (for $F/D$, random subspaces of equal rank), so the dashed line marks equality. \textbf{(a)}~In every model, the band moves the logits less than random changes ($D$, \markD) but its prediction change ($K$, \markK) and loss change (\markL) are several times as large (95\% intervals, adjusted jointly for $D$, $K$ and $F/D$; grey lines join $D$ and $K$). Per unit of logit distance ($F/D$, \markF: the consequence spectrum's bottom-band value), the excess is larger still. \textbf{(b)}~Per document (medians and 10--90\% ranges), $D$ is below random in every document and $F/D$ above random in most. \textbf{(c)}~The pattern holds in almost every case at other positions, on code and in instruction-tuned models.}
\label{fig:overview}
\end{figure*}

\subsection{The consequence spectrum gives the prediction change per unit of logit distance}
\label{sec:change}
\label{sec:spectrum}

We change the final hidden state within one band $b$ by replacing its component in that band with the calibration mean $\mu$. This mean is the average of $u$ over a separate set of calibration documents (\cref{sec:reversal}). As references, we apply the same replacement within each of eight random subspaces. Each is drawn uniformly from the state's whole space and has as many directions as the band:
\begin{equation}
  \delta u = -\,P_b\,(u - \mu), \qquad \delta u^{\mathrm{rand}}_j = -\,Q_j\,(u - \mu).
  \label{eq:change}
\end{equation}
Here $P_b = \sum_{i \in b} v_i v_i^\top$ is the orthogonal projector onto band $b$, and $Q_j$ is the projector onto the $j$th random subspace. In words, every document keeps the same, average component in the band and loses only whatever distinguished it there, while its components outside the band are untouched.

The \emph{consequence spectrum} gives, for each band, the prediction change per unit of logit distance relative to that of random changes. Here we measure the prediction change by the Fisher term $F$ rather than by $K$. This choice makes $F/D$ independent of the change's size, because $F$, like $D$, scales with the square of the change. We pool both over documents and divide by the same ratio for the random references:
\begin{equation}
  \boxed{\begin{gathered}\textbf{Consequence spectrum}\\[2pt] C_b = \frac{\sum_x F_{xb} \,/\, \sum_x D_{xb}}{\sum_x \overline{F}^{\mathrm{rand}}_{xb} \,/\, \sum_x \overline{D}^{\mathrm{rand}}_{xb}}\end{gathered}}\,.
  \label{eq:perunit}
\end{equation}
Here $x$ indexes the documents, and a bar denotes the mean over the eight references. A value of one means that the band's changes move the prediction as much per unit of logit distance as random changes do.

For a single singular direction, the singular value drops out of the ratio $F/D$. Specifically, a change $\epsilon v_i$ moves the logits by $\delta\ell = \epsilon\,\sigma_i a_i$ (\cref{eq:svd}), so
\begin{equation}
  F/D = \tfrac12\,\epsilon^2\sigma_i^2 \operatorname{Var}_p(a_i) \,\big/\, \epsilon^2\sigma_i^2 = \tfrac12 \operatorname{Var}_p(a_i),
  \label{eq:cancel}
\end{equation}
for any $\sigma_i > 0$. For a band, a common factor on its singular values cancels in the same way, so $C_b$ does not depend on their overall size. A direction can therefore have a small singular value and still a large prediction change per unit of logit distance. Its singular value is small when the vocabulary as a whole varies little along it, whereas its prediction change is large when the few probable tokens differ along it. \Cref{app:measurement} explains why the references of $C_b$ are not rescaled to equal length.

\section{The bottom band changes the prediction disproportionately to its logit distance}
\label{sec:reversal}

To test the hypothesis across model sizes and families, we evaluate ten base models from three families: seven Pythia models from 70M to 6.9B parameters \citep{biderman2023pythia}, Llama-3.1-8B \citep{grattafiori2024llama} and two Qwen3 base models of 1.7B and 8B parameters \citep{yang2025qwen3}. Each model reads 1{,}536 documents of 128 tokens from OpenWebText \citep{Gokaslan2019OpenWeb}. We take its final hidden state at the 128th token, from which it predicts the next token. Of these documents, 512 serve as calibration documents, which give the calibration mean $\mu$. The other 1{,}024 measure the effects. For every comparison, intervals are 95\% bootstrap intervals over documents unless stated otherwise. \Cref{app:measurement} describes the models, documents and intervals in full.

\subsection{At equal length, the bottom band moves the logits less but changes the prediction more}
\label{sec:equal}

To test whether changes that move the logits less can still change the prediction more, we compare each document's bottom-band change with its random references at equal length. To this end, we rescale each random change, document by document, to the length of the band's change. The prediction change, however, does not say whether the new prediction is better or worse. We therefore also measure the loss change, the change in the negative log probability of the observed next token $y$, which training minimizes:
\begin{equation}
  \Delta\mathrm{NLL} = \log p_y - \log p'_y.
  \label{eq:nll}
\end{equation}
For each measure, we report the \emph{ratio to random}: the band's mean over documents divided by the mean of its references. For a single document, we use instead the \emph{loss excess}, the band's loss change minus the mean loss change of its references.

The bottom band's changes move the logits less than their references, as their small singular values imply. Nevertheless, in every model, their prediction change and their loss change are several times as large (\cref{fig:overview}a).

The two measures therefore disagree along these directions. Counted over all tokens equally, this change of the logits is small. Weighted by the probabilities of the tokens it moves, as in the prediction change, it is large. As a result, a method that ranks changes by how far they move the logits places these changes below random changes that matter less for the prediction.

The average effect is not spread evenly over documents (\cref{fig:overview}b). Per unit of logit distance, the bottom band exceeds its references in 83 to 93\% of documents. Its total prediction change, however, is larger in only 47 to 68\% of them, because the band's change moves the logits several times less. Moreover, the tenth of documents with the largest loss excess carries most of the total loss excess, and \cref{sec:candidates} identifies these documents.

The prediction change and the loss change stay above those of the references in three further comparisons. First, the references are rescaled on average or not at all, instead of document by document. Second, the normalization's scale is recomputed after the change, as in the model itself, rather than held fixed. We call this the \emph{natural path}. Third, the 16 coordinates of the state with the largest mean square or variance on the calibration documents are projected out. In addition, the excess is not a property of directions of large variance. Compared in the same way, a subspace of equal size along the states' directions of largest variance shows a far smaller excess per unit of logit distance than the bottom band. \Cref{app:robust-equal,app:band-consequences} report these checks, the per-document distributions and the change of the most probable token.

The closest observation is Figure~8 of \citet{cancedda2024spectral}: removing this band after the final layers raises the loss more than removing a random subspace of the same rank. Our comparison adds changes of equal length, so that the band's larger effect cannot come from a larger change. It also separates the logit distance from the prediction change and tests the reason (\cref{sec:why}).

\begin{figure*}[!t]
\centering
\includegraphics[width=\textwidth]{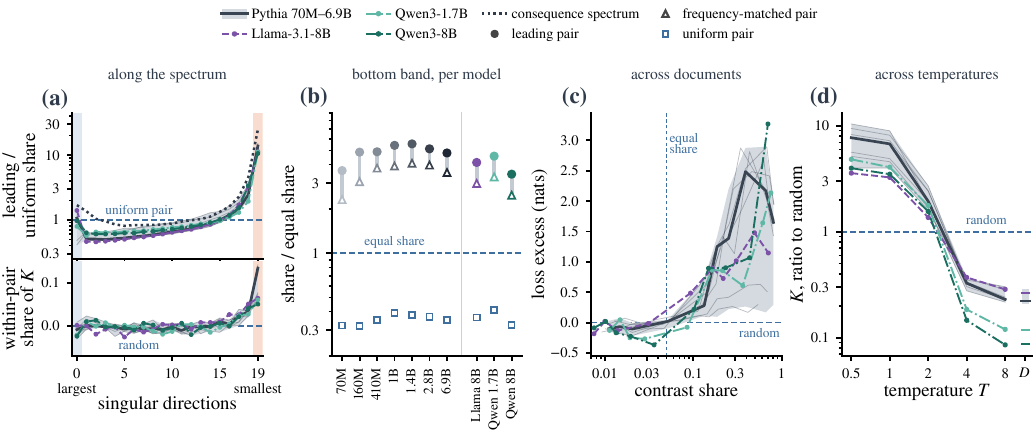}
\caption{\textbf{The bottom band carries a large share of the difference between the leading candidates, and its prediction change falls on them.} Lines are single models; for the seven Pythia models, the shaded region spans their range and the bold line is their median. \textbf{(a)}~Per band, the leading pair's contrast share relative to a uniform pair's rises over the last bands together with the consequence spectrum (dotted; Pythia median), and so does the share of $K$ within the leading pair (lower panel; band minus random). \textbf{(b)}~At the bottom band, the leading pair's share is several times the band's share of the directions, more than a frequency-matched pair's and far above a uniform pair's, which lies below the band's share of the directions. \textbf{(c)}~The loss excess rises once a document decile's mean share exceeds the band's share of the directions (vertical line). \textbf{(d)}~The $K$ ratio to random falls below one as a higher softmax temperature flattens the prediction, toward the ratio of the logit distances (right edge, $D$).}
\label{fig:mechanism}
\end{figure*}

\subsection{Along the spectrum, only the last bands change the prediction disproportionately}
\label{sec:cbspectrum}

To locate the effect along the spectrum, we compute the consequence spectrum over all twenty bands (\cref{fig:teaser}b). Across most of the spectrum, the prediction change per unit of logit distance stays within about a factor of two of that of random changes. The large excess appears only over the last few bands, and the bottom band has the largest $C_b$ in every model.

This rise does not follow from the small singular values themselves, because their overall size cancels from $C_b$ (\cref{sec:spectrum}). Instead, by \cref{eq:cancel}, the rise reflects how the probable tokens differ along these directions. \Cref{sec:why} tests whether the leading candidates differ most along the bottom band.

Ranking directions by singular value therefore places last the directions whose changes matter most per unit of logit distance. \Cref{app:occupancy} shows the singular values across the bands.

\subsection{The comparison holds at other positions, on code and in instruction-tuned models}
\label{sec:generality}

To test whether the result generalizes, we repeat the evaluation under three kinds of conditions. First, evaluating at the 8th and 32nd token of web text varies the amount of context. Second, code at the 8th, 32nd and 128th token varies the domain of the text. These code documents are 1{,}536 correct solutions from the training split of DeepMind Code Contests, one C++, Java or Python solution per problem \citep{li2022competition}. Third, two instruction-tuned models, Llama-3.1-8B-Instruct and the post-trained release of Qwen3-8B, vary the training stage of the weights. These models read web text at the same three positions, without a chat template.

The bottom band's prediction change per unit of logit distance stays well above random in all 56 model--condition--position combinations (\cref{fig:overview}c). The consequence spectrum, moreover, peaks at the bottom band in every one of them. The comparison at equal length also holds in almost all of them. \Cref{app:generality} reports every comparison by condition and position.

\section{The bottom band carries a large share of the difference between the leading candidates}
\label{sec:why}

\Cref{sec:reversal} found that the bottom band changes the prediction disproportionately to its logit distance. By \cref{eq:cancel}, this can happen when the few probable tokens differ along a direction although the vocabulary as a whole varies little along it. Among these tokens, the pair form of the Fisher term (\cref{eq:pairs}) weights each pair by the product of their probabilities. The largest weight therefore falls on the two most probable tokens, the leading candidates. This section tests three predictions of this account, one in each subsection.

\subsection{The leading pair's share follows the consequence spectrum and marks where the loss rises}
\label{sec:candidates}

To test whether the leading candidates differ along the bottom band, we measure how much of their difference lies in it. For each document, let $w_1$ and $w_2$ be the unembedding vectors of the two most probable tokens of the unchanged prediction. A change $\delta u$ moves their log-odds by $(w_1 - w_2)^\top \delta u$, and we call this difference the leading pair's \emph{contrast}. A band's \emph{contrast share}, in turn, is the fraction of the contrast's squared length that lies within the band, $\|P_b(w_1 - w_2)\|^2 / \|w_1 - w_2\|^2$. For a direction drawn uniformly at random, the expected share equals the band's share of the directions, 5\%. This share bounds the leading pair's term in the pair form of $F$ (\cref{eq:pairs}): because a band change lies within the band, it moves the pair's log-odds by at most $\|P_b(w_1 - w_2)\|\,\|\delta u\|$. A large share therefore makes a large contribution possible.

Along the whole spectrum, we compare this share with that of two tokens drawn uniformly from the vocabulary. Because $\sigma_i^2$ sums the tokens' squared coordinates along $v_i$ (\cref{sec:unembedding}), this uniform pair's expected share of a band is the band's share of the squared singular values. Dividing by it therefore removes the band's singular values, as the random references do in $C_b$. The ratio of the two shares then measures how much more the leading candidates differ along a band than the vocabulary as a whole. As the account predicts, this ratio follows the consequence spectrum (\cref{fig:mechanism}a): it stays near or below one for most bands and rises steeply over the last few. Moreover, the logarithms of the two spectra correlate at 0.95 to 0.98 across the ten models.

At the bottom band itself, the leading pair's share is several times 5\%, whereas the uniform pair's share is below it, as the band's small singular values imply (\cref{fig:mechanism}b). The high share could, however, reflect only that the leading candidates tend to be frequent tokens. To test this, we use a frequency-matched pair, two distinct tokens drawn in proportion to their mean probabilities on the calibration documents. This pair already has many times the uniform pair's share (triangles). The leading pair's share nevertheless exceeds the frequency-matched pair's in all ten models. The high share therefore has two parts. Most of it is common to probable tokens, as the account allows, and the candidates of a given context add a further, smaller part.

Because the share bounds how far a band change can move the leading pair's log-odds, the loss excess should concentrate in documents with a large share. The data follow this prediction (\cref{fig:mechanism}c). Specifically, among the deciles of documents sorted by share, the loss excess stays near or below zero in every model while the decile's mean share is below 5\%. It rises only above that level. The per-document prediction change likewise rises with the share. This concentration identifies the documents in which the total effect of \cref{sec:equal} is large: the average loss excess is large although the median document shows little. \Cref{app:robust-candidates} gives the regression of the loss excess, the frequency-matched comparisons and the per-document relations, and \cref{app:generality} the range of the share under each condition.

\begin{figure*}[!t]
\centering
\includegraphics[width=0.872\textwidth]{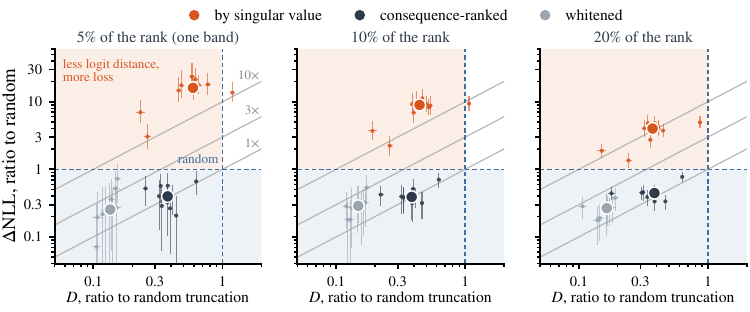}
\caption{\textbf{Truncating the unembedding by singular value moves the logits less than random truncation but increases the loss more. Ranking the same directions by their expected Fisher term avoids this cost.} Each panel shows one share of the rank removed. Each truncation is placed by its logit distance $D$ (horizontal) and its loss change $\Delta$NLL (vertical), both relative to random truncation of the same rank, so that the dashed lines mark equality. Small points are models, with their bootstrap intervals; large points are medians over the models. The warm region holds truncations that move the logits less but increase the loss more than random truncation. The grey diagonals mark equal loss per unit of logit distance, at one, three and ten times that of random truncation. The three truncations remove the directions with the smallest singular values, the singular directions with the smallest expected Fisher term (consequence-ranked), or the subspace that the reconstruction error on the states selects (whitened, as in SVD-LLM).}
\label{fig:compression}
\end{figure*}

\subsection{The prediction change falls on the leading candidates}
\label{sec:falls}

The contrast share shows where the leading candidates differ, but not whether the prediction change caused by a band change falls on them. To test this, we split the prediction change by the exact chain rule of the KL divergence into three parts: a change in the pair's total probability against the rest, redistribution within the leading pair, and redistribution among the remaining tokens. The first two parts together account for 85 to 99\% of the band's excess prediction change over its references. The excess therefore falls largely on the leading candidates. Of the three parts, the one within the pair depends on the change only through the pair's log-odds. This part also follows the consequence spectrum (\cref{fig:mechanism}a, lower panel). Its share of the band's prediction change, minus the same share for random changes, stays near zero for most bands and rises over the last few.

If the state's bottom-band component carries the choice between a context's candidates, replacing a document's bottom-band component with another document's should move its prediction toward the donor's. We test this with donors whose top token is the recipient's runner-up, so that the replacement can favour one of the recipient's own candidates. In almost every model, the replacement shifts the recipient's leading-pair log-odds toward the donor's more than random replacements of the same length do. \Cref{app:robust-candidates} gives the split of the prediction change and the donor replacements, including donors drawn at random.

\subsection{The excess needs a concentrated prediction}
\label{sec:concentrated}

The pair form also predicts that the excess depends on the concentration of the prediction. Raising the softmax temperature $T$ within each document makes the prediction more uniform and spreads the Fisher term's weight over more pairs. In the high-temperature limit, the prediction change becomes proportional to the logit distance (\cref{app:geometry}), so the band's ratio to random for $K$ must fall toward that for $D$. The data show where this fall occurs (\cref{fig:mechanism}d): the band's $K$ ratio stays above one at moderate temperatures and falls below one at high ones, toward the ratio of the logit distances. Across documents, likewise, the loss excess falls with the entropy of the unchanged prediction in every model. \Cref{app:robust-candidates} gives the ratios at the lowest and highest temperatures and the entropy relation.

Together, the three tests support the account stated at the start of this section: the bottom band carries a large share of the difference between the leading candidates, the band's prediction change falls on them, and the excess fades as the prediction is made uniform.

\section{Consequences for distance-based methods}
\label{sec:consequences}

Methods that interpret, edit or compress language models often judge a direction, or a change along it, by how far it moves the representation or the logits. This practice treats changes of equal distance as equally consequential for the prediction. The consequence spectrum, however, shows that this equivalence fails most at the bottom band (\cref{sec:cbspectrum}). Distance-based methods therefore undervalue the directions that matter most per unit of logit distance, and rankings by singular value place them last. Compression allows a direct test of this consequence, because the loss can be measured after truncation (\cref{sec:compression}).

\subsection{In compression, truncation by singular value increases the loss more than random truncation}
\label{sec:compression}

At the unembedding, the ranking by singular value that low-rank compression uses can be tested directly. Specifically, replacing the removed directions' component with the calibration mean, as in \cref{eq:change}, equals truncating the unembedding to the other directions with a corrected bias. The band change's logit distance and loss change are therefore those of the truncation. We evaluate these truncations without retraining.

At 5, 10 and 20\% of the rank, we compare three truncations with random truncation of the same rank. First, \emph{truncation by singular value} removes the directions with the smallest singular values, which at 20\% are the four bottom bands. Second, \emph{consequence-ranked truncation} ranks the same singular directions by their expected Fisher term of removal on the calibration documents and removes the smallest. This ranking follows Fisher-weighted criteria \citep{hsu2022fwsvd}. Third, \emph{whitened truncation} adapts SVD-LLM, which minimizes the logit reconstruction error on the calibration documents \citep{wang2024svdllm}. Unlike the other two, it may select any subspace of the given rank.

Truncation by singular value reproduces the result of \cref{sec:reversal} (\cref{fig:compression}): at every share of the rank, it increases the loss more than random truncation in every model, and it moves the logits less in almost every model. Changing only the ranking removes this cost. Specifically, consequence-ranked truncation removes a substantially different set of directions. It increases the loss less than random truncation in all ten models at 10 and 20\% of the rank, and in most at 5\%. A ranking computed once on the calibration documents thus suffices, because the per-unit excess holds in most documents (\cref{sec:equal}). In addition, whitened truncation increases the loss even less in most models. It weights each direction by the states' variance along it rather than by the singular value alone, and at 20\% of the rank it keeps part of the four bottom bands. A Fisher criterion is therefore not needed to avoid the cost. Even so, its loss change per unit of logit distance exceeds that of random truncation in almost every model. The cost therefore lies in ranking by the singular values alone, not in truncation as such. \Cref{app:robust-truncation} defines the three truncations and compares them in detail.

\subsection{Implications for interpretation, editing and representation comparison}
\label{sec:implications}

The same undervaluation bears on interpretation, editing and the comparison of representations. Interpretability work has described the bottom singular directions by their small effect on the logits \citep{elhage2021mathematical}, for example as a null space into which entropy neurons write \citep{stolfo2024confidence}. A ranking by logit distance would then place components that write into these directions below components that change the prediction less. Editing methods, including concept erasure, in turn take a small or null change of the representation as their safeguard against side effects \citep{fang2025alphaedit,belrose2023leace}. This safeguard would protect the prediction least along these directions. Similarity measures such as linear CKA compare representations by distances or variance-weighted correlations \citep{kornblith2019similarity}. Two states can then be close yet differ in their predictions. This is consistent with evidence that these measures can miss changes that alter a network's function \citep{ding2021grounding}. In each case, measuring the effect on the prediction, for example by the KL divergence or the loss change document by document, would register what distance misses. These implications follow from the measured result but are not tested on the methods themselves.

\section{Related work}
\label{sec:related}

\paragraph{Magnitude as a guide to importance.} Treating a component's size as its importance has a long history in compression. For example, optimal brain damage defined a parameter's importance as the increase in training error that deleting it causes, and showed that ranking by this quantity damages a network less than ranking by magnitude \citep{lecun1990optimal}. Low-rank compression likewise ranks directions by a magnitude, the singular value, which minimizes reconstruction error \citep{eckart1936approximation,wang2024svdllm}. By contrast, loss-aware criteria such as Fisher weighting weigh components by their estimated effect on the loss \citep{hsu2022fwsvd,chekalina2026scalable,shihab2026beyond}. Inside transformer blocks, moreover, the small singular components of weights can carry what a model has learned \citep{hsu2022fwsvd,staats2025small}. These studies show that importance need not follow the singular value inside the network. We measure the same divergence at the unembedding, where it can be traced to the tokens that a change moves (\cref{sec:why}).

\paragraph{Prediction as a guide to representations.} Interpretability has moved from reconstruction to prediction in judging what a representation keeps. For example, sparse autoencoders are evaluated by their effect on the loss \citep{gao2025scaling} or trained on the KL divergence of the output \citep{braun2024identifying}. Similarly, the tuned lens ranks directions by the KL divergence that their removal causes \citep{belrose2023tuned}, and the choice of metric changes which components look important \citep{zhang2024towards}. Our results locate where this choice matters most at the final hidden state: along the bottom band, assessments by logit distance and by prediction change disagree most.

\paragraph{Geometry of the prediction.} The Fisher metric describes local changes of a predictive distribution \citep{amari1998natural,martens2020new}, and information geometry has been applied to probing and steering representations that define softmax distributions \citep{park2026information}. Our two measures connect to this geometry: the logit distance is, up to vocabulary size, the squared length under the causal inner product of \citet{park2024linear}. The consequence spectrum, in turn, compares the two measures band by band, much as quotients of two metrics locate likelihood-informed subspaces in inverse problems \citep{cui2014likelihood,spantini2017goal}. \Cref{app:geometry} develops these connections.

\paragraph{The unembedding's bottom spectrum.} Prior studies have characterized the unembedding's bottom singular directions by their role inside the network or by their effect on the top prediction. Removing them at intermediate layers raises the loss through attention sinks \citep{cancedda2024spectral}, and entropy neurons write into them and act through the final normalization's scale \citep{stolfo2024confidence,baroni2026small}. After their removal at the final layer, top-token rankings (HARP; \citealp{hu2026harp}) and task accuracy \citep{joshi2025calibration} change little. In HARP's basis, applied to base Llama-3.1-8B on our documents, the same removal nevertheless changes the top token more often than equal-length random removals. HARP's evaluation differs in its prompts and scoring. To these measurements we add the prediction change and the loss change, against random references of equal length. With the normalization's scale held fixed, our measurement also isolates a second, direct route beside the route through the scale. \Cref{app:prior} compares the results with each of these studies and with further work on compression, interpretability and the softmax output.

\section{Discussion}
\label{sec:discussion}

The smallest singular directions of the unembedding matter disproportionately for next-token prediction, as the introduction hypothesized. The reason lies in two notions of size. One is the singular value, which measures how much the unembedding amplifies a direction over the whole vocabulary, with every token counted equally. The other is the prediction change, which responds to the differences among the few probable tokens. The bottom band is where these tokens differ while the vocabulary as a whole varies little (\cref{sec:why}). In the opening example, a change that barely moves the logits can therefore still decide between ``cup'' and ``glass''.

\paragraph{Amplification and consequence.} The singular value remains a reasonable guide over most of the spectrum, where the prediction change per unit of logit distance stays close to that of random changes (\cref{sec:cbspectrum}). The top band, moreover, matters in total through its large logit distance (\cref{app:band-consequences}). The singular value misjudges one class of directions: those along which the probable tokens differ more than the vocabulary as a whole. Along these directions, importance for the prediction has two parts. One is largely a property of the band: its prediction change per unit of logit distance exceeds random in most documents. The other is a property of the context: the share of the leading candidates' contrast that a document places in the band (\cref{sec:equal,sec:candidates}). The singular value, fixed by the weights alone, measures neither. Both parts also appear when the band's Fisher term is factored into the band's average sensitivity and the states' position within the band (\cref{app:robust-candidates}). Measuring importance on the prediction need not be costly, however. The expected Fisher term of removing each direction, computed once on calibration documents, ranks the directions by their effect on the prediction. In compression, this ranking avoids the cost of ranking by singular value (\cref{sec:compression}).

\paragraph{Two geometries.} Distance and variance give a representation one geometry, the same in every context. By contrast, the geometry of the prediction, given locally by the Fisher term, weights each direction by how the current probable tokens differ along it and therefore changes with the context. At the output, our results show that the two geometries disagree most along the directions that the unembedding amplifies least. Whether the same disagreement holds at earlier layers, where the prediction depends on a representation through further computation, remains to be tested.

\paragraph{A possible training origin.} In linear and unconstrained-feature models, the part of the output weights that excludes implausible tokens grows during training, while the part that sets the odds among plausible tokens stays finite \citep{thrampoulidis2024implicit,zhao2024implicit}. \Cref{app:prior} summarizes these results. If this separation maps onto the singular directions, it places the distinctions among plausible candidates at the bottom of the spectrum. Our data are consistent with the bottom half of this prediction: the share of $K$ within the leading pair exceeds that of random changes more in the bottom band than in the top band in all ten models. The other half, that the top directions carry the exclusion of implausible tokens, is not established (\cref{app:robust-candidates}). Testing this origin would require tracking the organization during training and comparing controlled training objectives.

\paragraph{Limitations.} We study one site, the final hidden state, in models up to 8B parameters, on raw web text and code up to the 128th token and without chat formatting for the instruction-tuned models. In addition, changes of equal length need not produce states that occur naturally. Similarly, the compression experiment truncates the unembedding without retraining. Likewise, the implications for interpretation, editing and representation comparison are not tested on the methods themselves.

\section*{Acknowledgements}
AI assistants were used to develop code and help with manuscript revision.

\bibliography{references}
\bibliographystyle{icml2026}

\newpage
\appendix
\crefalias{section}{appendix}
\crefalias{subsection}{appendix}
\onecolumn
\section{Computing the unembedding's singular directions and values}
\label{app:readout}

The singular directions and values depend only on the weights of the unembedding and of the final normalization. We compute them in six steps.
\begin{enumerate}
  \item We write the path from the final hidden state to the prediction as: final residual $r$; final LayerNorm (subtract the mean of the $d$ coordinates, divide by the scale $s(r)$, multiply coordinate $j$ by $\gamma_j$, add $\beta_j$); unembedding $W_{\mathrm{out}}$; softmax.
  \item We hold $s(r)$ at its unchanged value, which makes the map from a change of the normalized core to the change of the logits linear, $\delta\ell = W_{\mathrm{out}}\,\mathrm{diag}(\gamma)\,\delta z$.
  \item We remove two components that cannot change the prediction. First, the normalized core has mean zero, so we write $\delta z = R\,\delta u$ in an orthonormal (Helmert) basis $R$ of the $q = d-1$ dimensional mean-zero subspace. Second, a shared shift of all logits leaves the softmax unchanged, and $C_V = I - \mathbf 1\mathbf 1^\top / V$ removes this component.
  \item We form the unembedding as analysed here, $W = C_V W_{\mathrm{out}}\,\mathrm{diag}(\gamma)\,R$, of size $V \times q$ ($50{,}304 \times 2{,}047$ in Pythia-1B).
  \item We accumulate the $q \times q$ Gram matrix $W^\top W$ in float64 over blocks of vocabulary rows. Its eigenvectors are the right singular vectors of $W$ (the singular directions), and its eigenvalues are the squared singular values.
  \item We sort the directions by singular value and divide them into twenty bands of equal size, with band edges $\mathrm{round}(bq/20)$, $b = 0,\dots,20$, rounded half up.
\end{enumerate}
These steps give \cref{eq:readout} and the logits $\ell = W u + \ell_0 + c(r)\,\mathbf 1$, with $z = (r - \bar r\,\mathbf 1)/s(r)$ the normalized core and the normalized state $u = R^\top z$. There, $\ell_0$ collects the biases $\beta$ and $b_{\mathrm{out}}$, and $c(r)$ is a shift common to all tokens. In addition, all rows of the unembedding enter $W$, including reserved and padding rows, whereas the leading candidates of \cref{sec:why} are taken among the tokenizer's tokens. The padding rows, in turn, hold a mean probability below $3 \times 10^{-7}$ and are not the most probable token in any evaluated document. They therefore enter $D$ and the singular values but contribute little to $F$, $K$ and the loss change.

The number of directions is $q = 511$, $767$, $1{,}023$, $2{,}047$, $2{,}047$, $2{,}559$ and $4{,}095$ in the seven Pythia models (70M, 160M, 410M, 1B, 1.4B, 2.8B, 6.9B), whose bottom bands hold $26$, $38$, $51$, $102$, $102$, $128$ and $205$ directions.

\paragraph{RMSNorm models.} In Llama-3.1-8B and the two Qwen3 models the final normalization is an RMSNorm, $z = r / s(r)$ with $s(r) = (\|r\|^2/d + \epsilon)^{1/2}$, which subtracts no mean. Step 3 then removes only the shared logit component, $R$ is the identity, $q = d$ and $W = C_V W_{\mathrm{out}}\,\mathrm{diag}(\gamma)$: $q = 4{,}096$, $2{,}048$ and $4{,}096$ in Llama-3.1-8B, Qwen3-1.7B and Qwen3-8B, whose bottom bands hold $205$, $102$ and $205$ directions. Their unembeddings have $V = 128{,}256$ rows (Llama) and $151{,}936$ rows (Qwen). In the implementation, each RMSNorm model is rewritten exactly as a LayerNorm model of width $d + 1$, by a Helmert lift of the residual with unit weight and no biases. This rewriting changes no logit, and the singular directions are unchanged. The singular values share one common factor, which leaves every reported ratio unchanged, and the squared lengths acquire a common factor $(d+1)/d$, which we divide out.

\section{The two geometries of the prediction}
\label{app:geometry}

This appendix develops the logit distance and the Fisher term of \cref{sec:measures}. It shows that the logit distance is a distance between predictions, derives the high-temperature limit that joins the two measures, relates the prediction change to the loss change, and connects the consequence spectrum to the Gauss--Newton metric and to the causal inner product.

\paragraph{The logit distance as a distance between predictions.} The centred logits are the centred log-ratio transform of the predicted distribution. Therefore $D$ is exactly the squared Aitchison distance between $p$ and $p'$, the standard distance between compositions \citep{aitchison1982statistical}.

\paragraph{The high-temperature limit.} The prediction change and the logit distance are the two ends of one family indexed by the softmax temperature $T$. To second order, $T^2\,\mathrm{KL}(\mathrm{softmax}(\ell/T)\,\|\,\mathrm{softmax}((\ell+\delta\ell)/T)) = \tfrac12\operatorname{Var}_{p_T}(\delta\ell)$ with $p_T = \mathrm{softmax}(\ell/T)$. In the limit of large $T$, $p_T$ tends to the uniform distribution, under which the variance of the centred change is $D/V$. Hence $T^2\,\mathrm{KL} \to D/(2V)$, and matching logits is the high-temperature limit of matching predictions \citep{hinton2015distilling}. The band's ratio to random for $K$ therefore tends to its ratio for $D$ (\cref{sec:concentrated}).

\paragraph{The quotient of the two metrics.} Pulled back to the state, the Aitchison metric is $G = W^\top W$, which is proportional to the covariance of the unembedding vectors and has the singular directions as its principal axes. Similarly, the Fisher metric averaged over documents is $\bar F = \mathbb{E}_x[W^\top \Sigma_{p(x)} W]$ with $\Sigma_p = \mathrm{diag}(p) - pp^\top$. For the softmax, $W^\top \Sigma_p W$ equals the Gauss--Newton matrix of the output \citep{martens2020new}. Along a fixed direction $v$, the quotient of the two metrics
\[
  \rho(v) = \frac{v^\top \bar F\, v}{v^\top G\, v}
\]
is twice the average Fisher term per unit of logit distance. It is the per-direction form of the consequence spectrum: for a singular direction, $\rho(v_i)$ is the document average of $\operatorname{Var}_{p(x)}(a_i)$ in \cref{eq:cancel}. Its extrema on the non-null subspace are the generalized eigenvectors of $(\bar F, G)$, as for the quotients of two metrics that locate likelihood-informed subspaces in Bayesian inverse problems \citep{cui2014likelihood,spantini2017goal}.

\paragraph{The prediction change and the loss change.} The two differ by a signed first-order term,
\begin{equation}
  \Delta\mathrm{NLL} = K + (p - e_y)^\top \delta\ell,
  \label{eq:lossdecomp}
\end{equation}
where $e_y$ marks the observed token $y$. This term averages to zero when the observed token is drawn from the prediction itself. On a given document, however, it takes either sign, so a document's loss excess can be negative.

\paragraph{Relation to the causal inner product.} The causal inner product of \citet{park2024linear} is $\gamma^\top \mathrm{Cov}(\gamma_y)^{-1}\gamma'$ on the unembedding vectors $\gamma_y$ of a uniformly drawn token, and the state carries the dual inner product, $\mathrm{Cov}(\gamma_y)$. Here our unembedding vectors $w_t$ take the role of the $\gamma_y$, so $\mathrm{Cov}(\gamma_y) = G/V$. A change $\delta u$ therefore has the squared length
\[
  \delta u^\top \mathrm{Cov}(\gamma_y)\,\delta u = \delta u^\top G\,\delta u / V = D/V .
\]
Up to vocabulary size, the logit distance is thus the squared length under the causal inner product, and $\rho$ compares this uniform weighting of tokens with the weighting by the model's prediction.

\section{Measurement details}
\label{app:measurement}

This appendix collects the measurement details that \cref{sec:instrument,sec:reversal} summarize: the notation, the models and documents, the reference scalings and the pooling, donor replacement, the natural path and the inference.

\begin{table}[t]
\caption{The paper's notation: the quantities from the final hidden state to the prediction, followed by the comparisons built on them.}
\label{tab:notation}
\centering
\small
\begin{tabular}{@{}lp{0.72\linewidth}@{}}
\toprule
Symbol or name & Meaning \\
\midrule
$r$ & the final hidden state: the residual stream at the last input position after the last block \\
$u$ & the normalized final hidden state, in the coordinates that the unembedding reads \\
$W$ & the unembedding: the unembedding matrix with the final normalization's weight folded in, as one centred linear map (\cref{eq:readout}) \\
$\sigma_i$, $v_i$, $a_i$ & a singular value of $W$, its singular direction and the pattern of logit changes along it (\cref{eq:svd}) \\
band $b$ & one of twenty groups of singular directions of equal size, $b = 0, \dots, 19$, from the largest singular values to the smallest; the bottom band ($b = 19$) holds the 5\% with the smallest singular values \\
$P_b$, $Q_j$ & the orthogonal projectors onto band $b$ and onto the $j$th of eight random subspaces of the same size \\
$\mu$ & the calibration mean of $u$ \\
$\delta u = -P_b(u - \mu)$ & the band's change: the document's component in the band replaced by the calibration mean's \\
$E = \|\delta u\|^2$ & squared length of the change; equal-length references match it on every document \\
$D = \|\delta\ell\|^2$ & the logit distance: squared length of the centred change of the logits $\delta\ell = W\delta u$, every token pair weighted equally \\
$F = \tfrac12\operatorname{Var}_p(\delta\ell)$ & the Fisher term: the second-order approximation to KL, with probability-weighted token pairs \\
$K = \mathrm{KL}(p\,\|\,p')$ & the prediction change: the KL divergence from the unchanged prediction $p$ to the changed prediction $p'$ \\
$\Delta\mathrm{NLL}$ & the loss change: the change in the negative log probability of the observed next token $y$ (\cref{eq:nll}); the \emph{loss excess} subtracts the references' mean loss change from the band's \\
ratio to random & the band's mean of a measure over documents divided by its references' mean; the references have equal length unless stated otherwise (\cref{sec:equal}) \\
$C_b$ & the consequence spectrum: band $b$'s pooled $F/D$ relative to unrescaled same-rank references (\cref{eq:perunit}); $C_{19}$ is the bottom band's value \\
leading pair & the two most probable tokens of the unchanged prediction (the leading candidates), with unembedding vectors $w_1, w_2$; their contrast is $w_1 - w_2$ \\
contrast share & the fraction of a token pair's contrast, by squared length, that lies in a band: leading, frequency-matched or uniform pair \\
band enrichment, state alignment & the two factors of the band's Fisher ratio (\cref{eq:decomposition}): the band's average sensitivity relative to the whole space, and the actual change's sensitivity relative to random directions within the band \\
parts of $K$ at the leading pair & the three terms of the exact chain rule of $K$: the change in the pair's total probability against the rest, the redistribution within the pair and the redistribution among the rest \\
\bottomrule
\end{tabular}
\end{table}

\paragraph{Models and documents.} The ten base models are the seven final Pythia checkpoints (70M, 160M, 410M, 1B, 1.4B, 2.8B and 6.9B parameters), Llama-3.1-8B, Qwen3-1.7B-Base and Qwen3-8B-Base. All are loaded at pinned revisions of their official weights and run in float32, although the Llama and Qwen weights are stored in bfloat16. Their documents come from rows 300{,}000 to 399{,}999 of OpenWebText. A hash of each document's prefix assigns it to one of three blocks of 512 documents: the calibration block and two confirmation blocks, on which the effects are measured. In addition, texts read by earlier analyses of the Pythia models are excluded by text and prefix hash. The Llama and Qwen models use the same interval and rules with their own tokenizers, so their documents differ from those of the Pythia models, of which about 2\% are shared. Finally, the Llama inputs are preceded by the model's beginning-of-text token.

\paragraph{Reference scalings and pooling.} The equal-length comparison of \cref{sec:equal} rescales each reference, document by document, to the squared length $E_{xb}$ of the band's change. Specifically, the $j$th reference is multiplied by $\lambda_{xbj} = \sqrt{E_{xb}/E^{\mathrm{rand}}_{xbj}}$, where $E^{\mathrm{rand}}_{xbj}$ is its own squared length. Two further scalings serve as checks. Calibration matching uses one coefficient per random subspace, chosen to match the band's mean squared length on the calibration documents. Unrescaled references are the replacements of \cref{eq:change} as they are. The consequence spectrum, however, keeps the unrescaled references. In each document, rescaling multiplies a reference's $F$ and $D$ by $\lambda_{xbj}^2$, so the document's ratio is unchanged. The pooled ratio, however, is a $D$-weighted average of the documents' ratios, and document-specific rescaling changes these weights. With rescaled references, the denominator of \cref{eq:perunit} would become
\[
 \frac{\sum_{x,j}\lambda_{xbj}^2 F^{\mathrm{rand}}_{xbj}}
      {\sum_{x,j}\lambda_{xbj}^2 D^{\mathrm{rand}}_{xbj}},
\]
which need not equal the unrescaled one.

\paragraph{Donor replacement.} In donor replacement, each document's band component is replaced by that of a partner document rather than by the calibration mean, $\delta u_i = -P_b(u_i - u_{j(i)})$. Documents are paired at random within each confirmation block. The matched-donor test of \cref{sec:falls} instead pairs calibration documents, because the confirmation blocks had already been read under random pairing.

\paragraph{The natural path.} On the natural path, the change is added to the final hidden state, $r' = r + s(r)\,R\,\delta u$, with $s(r)$ the normalization's scale (\cref{app:readout}), and the normalization is recomputed as in the model. This recomputed scale multiplies the changed state $u + \delta u$, and with it the state's contribution to the logits, by $s(r)/s(r')$. This factor can change the prediction's confidence in addition to the band's own effect. Because $s(r)$ is common to all changes of a document, changes of equal length in $u$ also have equal length in $r$.

\paragraph{Inference.} The document is the resampling unit, and in donor replacement the donor pair. Each interval comes from 20{,}000 paired bootstrap draws, stratified by confirmation block. Intervals of prespecified comparisons are 95\% intervals Bonferroni-adjusted within their family, and the other intervals are pointwise 95\% intervals. For a directional statement, the interval and the intervals from each half of the draws must lie on the claimed side of the null value. The point estimates of both confirmation blocks must also lie on that side, as must those of both prespecified halves of the eight random subspaces where these enter. A composite statement, such as $D$ below and $K$ above random, requires every component. A prespecified comparison that misses this rule is reported as not established. No numerical threshold beyond the null values was set.

\paragraph{Evidence classes.} Results in the appendices are labelled by how they were obtained. First, a \emph{prespecified} result is a comparison fixed in a dated contract before its outcome was read, and decided by the rule above. Its contract also lists any related outcomes read before it was fixed. Second, a \emph{descriptive} result is a quantity that the fixed analysis computes and reports without a test. Finally, a result \emph{computed after the prespecified results were known} is an analysis chosen after these results and is reported as an observation.

\paragraph{Verification.} Each run's results are recomputed by an auditor that shares no computation code with its producer and imports only a frozen registry of the protocol.

\section{Singular values and late-state variance across the bands}
\label{app:occupancy}

\begin{figure}[tb]
\centering
\includegraphics[width=\textwidth]{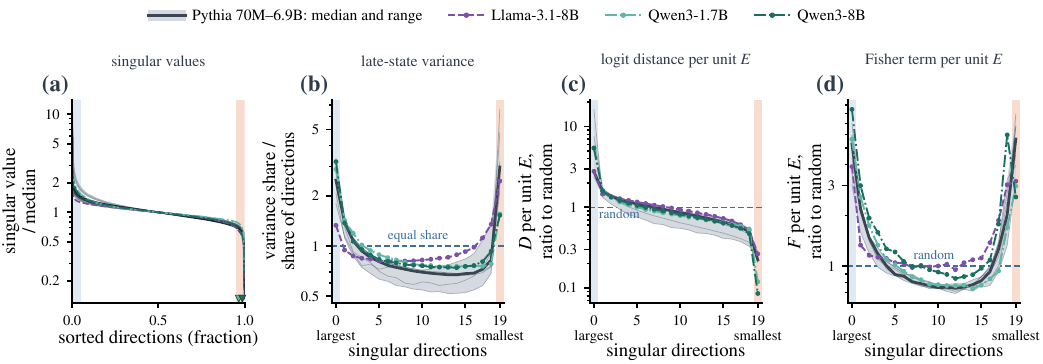}
\caption{\textbf{The singular values decline smoothly into the bottom band, whereas the late-state variance concentrates at both ends of the spectrum.} \textbf{(a)}~Sorted singular values divided by each model's median, one line per Llama or Qwen model and, for Pythia, the median and range of the seven models, with the top and bottom bands shaded; triangles mark Qwen's near-zero singular values below the plotted range. \textbf{(b)}~Each band's share of the late-state variance divided by its share of the directions: the bottom and top bands hold more than their share. \textbf{(c)}~The logit distance per unit of squared length $E = \|\delta u\|^2$, relative to random subspaces of the same rank, falls along the spectrum, whereas \textbf{(d)}~the Fisher term per unit $E$ rises above random at both ends. Their quotient is the consequence spectrum.}
\label{fig:gainaxis}
\end{figure}

This appendix reports how the singular values and the late-state variance, the variance of the states at the two sites before the final hidden state, are distributed across the bands (\cref{fig:gainaxis}). Specifically, the singular values change continuously at the bottom band's edge: the band's largest is within 0.3\% of the one just outside it in every model (\cref{fig:gainaxis}a). Only the Qwen bottom bands contain near-zero singular values, two in Qwen3-8B and three in Qwen3-1.7B, and \cref{app:robust-equal} compares the band with and without them. The bottom band's singular values are small but not negligible: their mean square is 0.31 to 0.41 of the average over all directions.

Panels (c) and (d) separate $C_b$ into its two factors, each per unit of squared length $E$ of the change and relative to random subspaces of the same rank. In this decomposition, the logit distance per unit $E$ falls along the spectrum, as the singular values do, whereas the Fisher term per unit $E$ is lowest over the middle bands and rises above random at both ends. Their quotient, the consequence spectrum, therefore rises from the top band to the bottom one: a line fitted to $\log C_b$ across the bands has a positive slope in every model (prespecified).

The late-state variance is measured at the residual stream after blocks $L-2$ and $L-1$ of $L$, on the same documents as the main evaluation. At both sites, the states are normalized by the final normalization and expressed in the unembedding's coordinates. Across the two sites, the bottom band holds on average 1.5 to 6.6 times its share of the directions in variance and the top band 1.3 to 3.4 times (\cref{fig:gainaxis}b; descriptive). The excess fades within one or two bands of each end, and the bands in between hold about their share or less. A prespecified comparison, moreover, tests the bottom band's excess directly, as its share of the variance minus its share of the directions, averaged over the two sites. This excess is above zero in every model (Bonferroni 95\% intervals, 20{,}000 draws). In the Pythia models it also stays above zero after the 16 coordinates with the largest calibration mean square, or the largest calibration variance, are projected out with the unembedding and the bands held fixed. \Cref{app:robust-equal} applies the same removal to the equal-length comparison in all ten models. This unequal variance across the bands is why the main comparison matches the length of the change rather than only its rank (\cref{sec:equal}).

\section{Robustness checks and additional analyses}
\label{app:robust}

This appendix reports the robustness of the equal-length comparison of \cref{sec:reversal} (\cref{app:robust-equal}) and further consequences of the band changes (\cref{app:band-consequences}). It then gives further analyses of the leading candidates of \cref{sec:why} (\cref{app:robust-candidates}) and further comparisons of the truncations of \cref{sec:compression} (\cref{app:robust-truncation}).

\subsection{Robustness of the equal-length comparison}
\label{app:robust-equal}

\begin{figure}[tb]
\centering
\includegraphics[width=\textwidth]{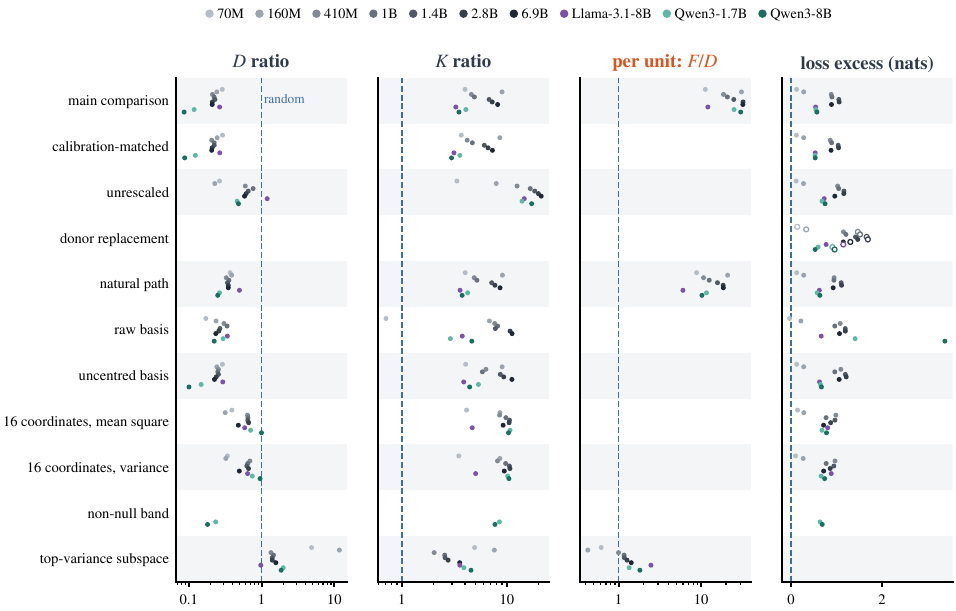}
\caption{\textbf{Under every alternative comparator and control, the bottom band's prediction change and loss excess stay above random, with one exception in Pythia-70M.} Rows give the main comparison (first row) and its alternatives; columns give the ratios to random of $D$ and $K$, the per-unit ratio $F/D$ and the loss excess. Points are point estimates, one per model; blank cells were not measured. $F/D$ does not depend on how the references are rescaled and is shown once for the first three rows; in the first row it is the consequence spectrum's bottom-band value $C_{19}$. On the natural path, $D$ and $F/D$ use calibration-matched references. In the donor row, filled points use calibration-matched and open points unrescaled references. The last row replaces the band by a subspace of equal rank along the states' directions of largest variance. Dashed lines mark random.}
\label{fig:controls}
\end{figure}

\Cref{fig:controls} repeats the comparison of \cref{sec:equal} under other scalings of the references, a donor as the replacement value, the natural path, two other bases, the removal of extreme coordinates and, in the Qwen models, without the band's near-zero directions. We say that the full result holds when $D$ is below random and $K$ and the loss excess are above random. In every model and under every alternative, $K$ and the loss excess stay above random, with one exception: in the raw basis, the $K$ ratio of Pythia-70M is below one and its loss excess is not above zero. The logit distance, by contrast, is not below random under some comparators, because their changes are shorter than the band's. Finally, a last paragraph compares the band with a subspace of the states' largest variance.

\paragraph{Reference scaling and donor replacement.} Against random changes rescaled by one calibration-matched coefficient, $D$ is below random, and the Fisher term, $K$ and the loss excess are above random in every model (prespecified). Against unrescaled random changes, likewise, the Fisher term, $K$ (3.3 to 21 times random) and the loss excess are above random in every model. $D$, however, is below random in only nine models. The reason is that the unrescaled references are shorter. In eight models, the band's change carries 2.8 to 5.6 times their squared length, and in Pythia-70M and Pythia-160M 0.92 and 0.96 times. In Llama-3.1-8B the factor is 4.5, so its $D$ is 1.19 times random (prespecified, not established), although per unit of squared length it is about a quarter of random. Finally, replacing the component with a donor's rather than with $\mu$ gives a positive loss excess in every model under both comparators.

\paragraph{The natural path.} On the natural path (\cref{app:measurement}), $K$ and the loss excess at equal length are above random in every model, and $D$ is below that of calibration-matched random changes. Against unrescaled random changes, however, the natural-path $D$ is not below random in the eight models from Pythia-410M up (prespecified, not established), while the natural-path Fisher term is above random in every model. Here, recomputing the scale multiplies all logits by a common factor, which acts like a change of temperature. We therefore remove from each change of the logits its component parallel to the unchanged logits. After this removal, the Fisher term is still above random in every model under both comparators. A uniform rescaling of the logits therefore does not fully account for the Fisher term's excess.

\paragraph{The band's non-null part in the Qwen models.} In the Qwen models a few coordinates on which the normalization weight $\gamma$ is close to zero give directions with near-zero singular values, below 0.05 of the bottom band's median: two in the bottom band of Qwen3-8B and three in that of Qwen3-1.7B. They carry 43\% (Qwen3-8B) and 38\% (Qwen3-1.7B) of the band change's squared length on average. We therefore repeat the comparison on the band's remaining directions, against random changes of the same squared length drawn outside the near-zero directions (prespecified). In this comparison, the full result holds and is stronger: $D$ is 0.18 and 0.23 times random, $K$ is 7.7 and 8.5 times random instead of 3.5 and 4.1, and the loss excess is 0.69 and 0.64 nats.

\paragraph{Alternative bases.} Defining the band in the raw unembedding basis reproduces the full result in every model except Pythia-70M, whose $K$ ratio is 0.71, below random, and whose loss excess is not above zero (prespecified in the Llama and Qwen models, descriptive in the Pythia models). In this basis, the loss excess of Qwen3-8B is 3.4 nats, the largest in \cref{fig:controls}. In the folded but uncentred basis, the full result holds in every model, with $D$ 0.10 to 0.29 times random and $K$ 3.9 to 11.2 times random.

\paragraph{Extreme-coordinate removal.} Projecting the 16 extreme coordinates, chosen on the calibration documents, out of both the change and the band leaves the full result in two of the three prespecified models and in all seven Pythia models. In Qwen3-8B only the logit comparison is not established: its $D$ ratios are 1.00 and 0.95, with intervals including one. Its $K$ ratio is 10.4 under both selections, and its loss excesses are 0.78 and 0.74 nats.

\paragraph{A top-variance subspace.} A subspace of the same rank along the states' leading principal directions moves the logits 1.3 to 11.7 times as much as random in nine models, and 0.98 times in Llama-3.1-8B (interval 0.95 to 1.00). Its $K$ is accordingly 2.0 to 7.6 times random. Per unit of logit distance, however, its $F/D$ ratio is 0.4 to 2.5, against 12 to 38 for the bottom band in the same run. The full result therefore holds only in Llama-3.1-8B, where this subspace's per-unit ratio, 2.5, is a fifth of the bottom band's 12.5 in the same run.

\subsection{Further consequences of the band changes}
\label{app:band-consequences}

\begin{figure}[tb]
\centering
\includegraphics[width=\textwidth]{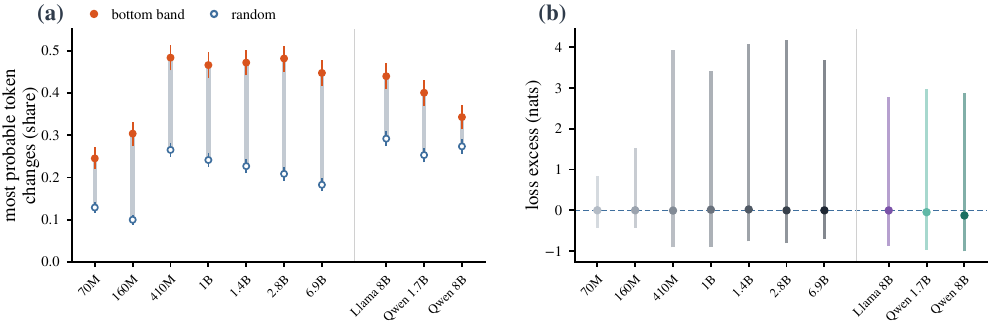}
\caption{\textbf{The bottom band's change replaces the most probable token more often than random changes, and its loss excess lies in the upper tail of documents.} \textbf{(a)}~Per model, the share of documents whose most probable token changes, under the band's change (filled) and under its random references of equal length (open), with 95\% intervals. \textbf{(b)}~Per model, the per-document loss excess: the median (dot) is near zero, and the 10--90\% range (bar) extends to several nats above it. The dashed line marks random.}
\label{fig:appconseq}
\end{figure}

\paragraph{Per-document distributions.} The per-document comparison of \cref{sec:equal} also holds with the exact $K$ in place of $F$: the band's $K$ per unit of logit distance exceeds that of its references in 77 to 92\% of documents, with a median per-document ratio of 4.9 to 12.1 (descriptive). The per-document loss excess, by contrast, has a median between $-0.13$ and $0.02$ nats in every model (\cref{fig:appconseq}b). Its upper tail carries the average: the tenth of documents with the largest excess carries 63 to 115\% of the summed excess. A share above 100\% means that the other documents together have a negative excess.

\paragraph{The most probable token.} The bottom band's change replaces the most probable token in 25 to 48\% of documents, against 10 to 29\% under its references (\cref{fig:appconseq}a). As a result, top-1 accuracy falls by 0.03 to 0.20, against at most 0.04 under the references (descriptive). When the most probable token is replaced, however, the new one is the unchanged prediction's runner-up less often than under random changes in eight of the ten models. This is consistent with the largest part of the split in \cref{sec:falls}, which moves probability between the leading pair and the remaining tokens.

\paragraph{The top band's total effect.} Removing the top band's component increases the loss as much as removing the bottom band's, or more, in nine of the ten models, by 0.96 to 1.75 nats against 0.17 to 1.23 nats. In Llama-3.1-8B it increases the loss less, by 0.60 against 0.79 nats. The top band's $C_b$, however, is 0.4 to 2.4, a small fraction of the bottom band's. Its total effect therefore comes from its large logit distance rather than from a large prediction change per unit of logit distance.

\subsection{Further analyses of the leading candidates}
\label{app:robust-candidates}

\paragraph{The frequency-matched pair.} The frequency-matched pair's share of the bottom band is 7.2 to 10.8 times the uniform pair's. The leading pair's share, in turn, exceeds the frequency-matched pair's by a factor of 1.29 to 1.59, with intervals above one in all ten models, and the share of the contrast between the first and third candidates exceeds it by a factor of 1.20 to 1.45 (descriptive). In addition, lexical similarity does not explain the share: the leading tokens differ only in case, a leading space or inflection in 126 of the 10{,}240 documents.

\begin{figure}[tb]
\centering
\includegraphics[width=\textwidth]{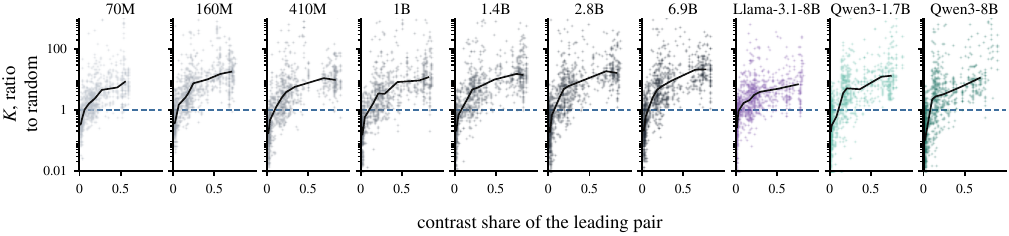}
\caption{\textbf{Larger leading-pair contrast shares go with larger prediction changes in each model.} Points show the per-document $K$ ratio against the bottom band's contrast share (1{,}024 documents per model), and lines the decile medians, which rise overall with the share in every model. The dashed line marks random, and the vertical axis is cut at 0.01 and 1{,}000.}
\label{fig:permodel}
\end{figure}

\paragraph{Document-level relations in each model.} \Cref{fig:permodel} shows the per-document relation of \cref{sec:candidates} in each model separately: the $K$ ratio rises with the bottom band's contrast share, with Spearman correlations of 0.72 to 0.82 (computed after the prespecified results were known).

\paragraph{Regression of the loss excess on share, entropy and margin.} In a prespecified regression across documents in the Llama and Qwen models, the loss excess rises with the contrast share and falls with the entropy of the unchanged prediction. The regression also includes the margin between the two leading logits: its slope is above zero in the two Qwen models and not established in Llama-3.1-8B. The seven Pythia models show the same entropy relation across thirds of documents: the loss excess is 0.23 to 2.02 nats in the lowest-entropy third, 0.14 to 1.08 nats in the middle third and 0.01 to 0.33 nats in the highest (descriptive).

\begin{figure}[tb]
\centering
\includegraphics[width=\textwidth]{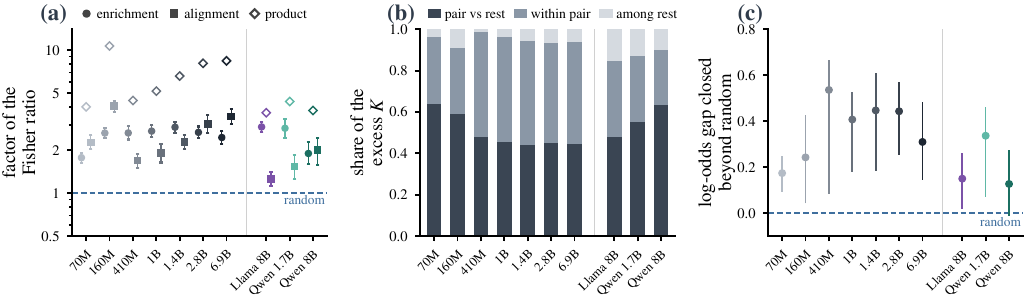}
\caption{\textbf{Both factors of the bottom band's Fisher ratio exceed one, its excess prediction change falls on the leading pair, and a matched donor moves the recipient's leading-pair log-odds toward its own.} \textbf{(a)}~Band enrichment (circles) and state alignment (squares), with 95\% intervals, and their product, the Fisher ratio (diamonds). \textbf{(b)}~The band's excess $K$ split by the chain rule of the KL divergence into the part between the leading pair and the rest, the part within the pair and the part among the remaining tokens; the first two parts carry most of it in every model. \textbf{(c)}~Per model, the median fraction of the recipient--donor log-odds gap that the bottom-band replacement closes beyond random replacements of the same length, for donors whose top token is the recipient's runner-up, with its interval. Dashed lines mark random.}
\label{fig:candidates}
\end{figure}

\paragraph{The chain rule at the leading pair.} The partition of \cref{sec:falls} describes which probabilities change, not which cause the loss. It puts 44 to 64\% of the band change's excess $K$ over its random changes in the part between the pair and the rest, 27 to 51\% in the part within the pair and 1 to 15\% in the part among the rest (descriptive; \cref{fig:candidates}b).

At the bottom band, the prespecified comparison is the within-pair share of the band change's $K$, as a ratio of sums over documents, minus the same share for its random changes. This quantity differs in its weighting from the band-by-band profile of \cref{fig:mechanism}a (lower panel). It is 0.13 to 0.27 in the Pythia models (descriptive), 0.10 in Llama-3.1-8B and 0.06 in Qwen3-1.7B. In Qwen3-8B it is 0.02, with an interval from $-0.04$ to $0.08$, so the comparison is not established there; its excess lies mainly in the part between the pair and the rest.

\paragraph{The between-pair-and-rest part across bands.} If the top directions carried the exclusion of implausible tokens, the top band's change would place a larger share of its $K$ in the part between the leading pair and the rest than the bottom band's. The data do not show this: this share is higher at both ends of the spectrum than over the middle bands, and the top band's exceeds the bottom band's in one prespecified model, Qwen3-8B, and in no Pythia model. The within-pair part, by contrast, exceeds that of random changes more at the bottom band than at the top band in all ten models (\cref{fig:mechanism}a, lower panel).

\paragraph{Random-donor log-odds transfer.} Replacing a document's bottom-band component with that of a randomly paired document makes the donor's top token the prediction more often than random replacements of the same length do. Random pairing does not, however, establish selective transfer between the recipient's leading candidates, because a random donor assigns them a median probability of only $10^{-6}$ to $10^{-4}$. To test it, we measure the fraction of the recipient--donor log-odds gap closed by bottom-band replacement beyond random replacements. Pooled over documents, this fraction is between $-0.05$ and $0.03$, with every interval including zero (prespecified in all ten models).

\paragraph{Matched-donor log-odds transfer.} The matched-donor test pairs each eligible calibration document with a donor whose top token is the recipient's runner-up, selected by a fixed hash. Its estimand is the median per-document fraction of the log-odds gap closed beyond random replacements, with intervals from 20{,}000 bootstrap draws over groups linked by shared donors. This median is positive in nine models, from 0.15 to 0.54 (prespecified). In Qwen3-8B, however, it is 0.13 $[-0.01, 0.27]$, so the comparison is not established there.

Four further descriptive results characterise the transfer. First, it does not exceed what the contrast geometry of \cref{sec:candidates} predicts, a transferred fraction equal to the band's contrast share minus the random changes' share: the median difference ranges from $-0.031$ to $0.006$, with no interval above zero. Second, the transfer depends on how the donor ranks the recipient's top token: it is 0.17 to 0.74 when the donor ranks that token 3rd to 100th, and 0.01 to 0.22 when it ranks it lower. Third, the donor's top token becomes the prediction in 48 to 66\% of documents, compared with 22 to 40\% under random replacements. Finally, restricting Qwen3-8B to the band's non-null part gives a median fraction of 0.15 $[0.04, 0.24]$.

\paragraph{Alternative temperatures and scoring rules.} The band's $K$ ratio to random falls from 3.6 to 10.4 at $T = 0.5$ to 0.09 to 0.30 at $T = 8$ (descriptive; \cref{fig:mechanism}d gives every temperature). At $T = 8$ it matches the ratio of the logit distances, 0.09 to 0.29, which it approaches as $T$ grows (\cref{app:geometry}). Under another proper scoring rule, the Brier divergence between the unchanged and the changed prediction \citep{gneiting2007strictly}, the band's change is likewise 2.6 to 14.9 times its random changes of the same length (descriptive).

\paragraph{The band's directions and the states' position.} A sensitive band does not by itself determine the effect of a change, because where the state lies within the band determines which directions the replacement moves. To separate the two, we compare the actual change with uniformly random directions of the same squared length, first within the band and then in the whole space. Let $F_{\mathrm{band}}$ be the actual Fisher term, $F_{\mathrm{rand}\in\mathrm{band}}$ its random-direction expectation within the band, and $F_{\mathrm{rand}}$ the expectation over the whole space, each averaged over documents. Unlike the eight random subspaces of \cref{sec:change}, these expectations average over all directions of the given squared length. Their ratio, the band's \emph{Fisher ratio}, factors exactly into a \emph{band enrichment}, the band's average sensitivity relative to the whole space, and a \emph{state alignment}, the actual change's sensitivity relative to random directions within the band:
\begin{equation}
  \frac{F_{\mathrm{band}}}{F_{\mathrm{rand}}} = \underbrace{\frac{F_{\mathrm{rand}\in\mathrm{band}}}{F_{\mathrm{rand}}}}_{\text{band enrichment}} \times \underbrace{\frac{F_{\mathrm{band}}}{F_{\mathrm{rand}\in\mathrm{band}}}}_{\text{state alignment}}.
  \label{eq:decomposition}
\end{equation}
For a uniformly random direction of squared length $E$ in a subspace of dimension $r$ with projector $P$, the expected Fisher term is $\tfrac{E}{2r}\operatorname{tr}\!\big(P\,W^\top \Sigma_p W P\big)$, with $\Sigma_p = \mathrm{diag}(p) - pp^\top$, so neither expectation requires sampling. Both factors exceed one in every model (\cref{fig:candidates}a). The band is therefore more sensitive on average, and the actual changes favour its more sensitive directions. Moreover, without the near-zero directions of the Qwen bands (\cref{app:robust-equal}), the state alignment of Qwen3-8B and Qwen3-1.7B is 4.2 and 3.1.

\subsection{Further comparisons of the truncations}
\label{app:robust-truncation}

\paragraph{The three truncations.} All three truncations replace the removed component with its calibration mean, $W[u - P(u - \mu)] + \ell_0 = W(I - P)u + (\ell_0 + WP\mu)$, where $P$ projects onto the removed directions and $\ell_0$ collects the biases. Consequence-ranked truncation removes, among the singular directions, those with the smallest expected Fisher term of removal on the calibration documents, $\kappa_i = \mathbb{E}_x\big[\tfrac12\,((u - \mu)^\top v_i)^2\, v_i^\top W^\top \Sigma_p W v_i\big]$, the curvature saliency of optimal brain damage \citep{lecun1990optimal} applied to directions. Whitened truncation factors the second moment of the calibration states about $\mu$, taken over all 128 positions of the calibration documents, as $M = SS^\top$. It uses $P = S V_d V_d^\top S^{-1}$, where $V_d$ holds the eigenvectors of $S^\top W^\top W S$ with the smallest eigenvalues. Among truncations of the given rank, this choice minimizes the mean logit reconstruction error on these states, as in SVD-LLM \citep{wang2024svdllm}.

\paragraph{Singular-value, random and consequence-ranked truncation.} At 20\% of the rank, truncation by singular value increases the loss by 0.11 to 1.24 nats more than random truncation. Consequence-ranked truncation increases it by 0.07 to 0.28 nats less than random truncation, and by 0.28 to 1.42 nats less than truncation by singular value. Their removal sets overlap by at most 27\% (descriptive).

The exceptions to the trends of \cref{sec:compression} are single models. First, truncation by singular value moves the logits more than random truncation in Llama-3.1-8B at 5 and 10\% of the rank, with $D$ ratios of 1.19 and 1.07. At 5\%, consequence-ranked truncation is not established below random truncation under the rule of \cref{app:measurement} in Pythia-6.9B, Llama-3.1-8B and Qwen3-1.7B, whose loss differences are $-0.02$ nats, although their plotted intervals lie below one. Finally, whitened truncation's loss change per unit of logit distance is below random truncation's only in Qwen3-8B at 5\% (descriptive).

\paragraph{Consequence-ranked and whitened truncation.} At 20\% of the rank, 47 to 63\% of the subspace that whitened truncation removes lies in the four bottom bands (descriptive). Whitened truncation also increases the loss less than consequence-ranked truncation in eight of the ten models at 5 and 10\% of the rank and in nine at 20\%. Accordingly, a per-document loss advantage of consequence-ranked truncation is established in no Llama or Qwen model, and in one Pythia model at 5\% (descriptive). At 20\%, consequence-ranked truncation's loss exceeds whitened truncation's by $-0.02$ to $0.17$ nats.

\paragraph{Fisher-whitened truncation.} Replacing $W^\top W$ in whitened truncation by the calibration mean of $W^\top \Sigma_p W$ fits the truncation to the Fisher term instead of the logit distance. This Fisher-whitened truncation increases the loss less than whitened truncation in one of the three prespecified models, Qwen3-1.7B at 5\% of the rank, and in none at 10 or 20\%. Among the Pythia models, it does so in two, three and four models at 5, 10 and 20\%, respectively (descriptive).

\section{Generality across positions, corpora and instruction-tuned models}
\label{app:generality}

This appendix reports the 56 model--condition--position combinations of \cref{sec:generality} (\cref{tab:generality}). Each condition and position has four prespecified comparison families. The first tests the equal-length comparison: the $D$ ratio below one, and the $K$ ratio, the loss excess and the band's own loss change, before the references are subtracted, above their nulls. The second tests the slope of $\log C_b$ across the bands. The third tests $C_{19}$ and its margin over the other bands, $C_{19} - \max_{b<19} C_b$, and the fourth the leading pair's contrast share against a uniform pair's. Intervals are Bonferroni 95\% within each family, from 20{,}000 draws. As in the main text, the comparisons of $C_{19}$ use the unrescaled references of \cref{eq:perunit}, whereas the first family uses references rescaled to the band's length. In all, every comparison holds in every combination except the equal-length comparison of Qwen3-8B at the 8th code token.

This exception lies in a cell with one of the lowest leading-pair shares, 5\%, and the band's near-zero directions carry 30\% of the band change's squared length there. The band's own loss change nevertheless remains above zero, and its consequence spectrum still peaks at the bottom band. More generally, the leading pair's share varies with the condition (descriptive). It is 18 to 28\% at the earlier web-text positions and in the instruction-tuned models and 9 to 18\% on code at the 32nd and 128th token. At the 8th code token it is 18 to 45\% in the Pythia models but 4 to 7\% in the Llama and Qwen models.

Finally, on the natural path, the band's $K$ ratio is 1.00 to 1.21 times its fixed-scale value, and its loss excess is 0.00 to 0.17 nats above the fixed-scale value, in all 56 combinations (descriptive).

\begin{table}[!htb]
\centering
\small
\caption{\textbf{The prespecified comparisons hold in all but one model--condition--position combination.} Each cell gives the number of models in which a prespecified comparison holds, with every interval on its side of the null: the equal-length comparison ($D$ below random; $K$ and the loss excess above random; the band's own loss change above zero), a positive slope of $\log C_b$ across the bands, $C_{19}$ above random and above every other band, and the leading pair's share of the bottom band above a uniform pair's.}
\label{tab:generality}
\begin{tabular}{@{}llccccl@{}}
\toprule
Condition & Token & Equal length & Slope & $C_{19}$ peak & Share & Exceptions \\
\midrule
Web text, base & 8th & 10/10 & 10/10 & 10/10 & 10/10 & -- \\
Web text, base & 32nd & 10/10 & 10/10 & 10/10 & 10/10 & -- \\
Code, base & 8th & 9/10 & 10/10 & 10/10 & 10/10 & Qwen3-8B ($K$, loss excess) \\
Code, base & 32nd & 10/10 & 10/10 & 10/10 & 10/10 & -- \\
Code, base & 128th & 10/10 & 10/10 & 10/10 & 10/10 & -- \\
Web text, instruction-tuned & 8th & 2/2 & 2/2 & 2/2 & 2/2 & -- \\
Web text, instruction-tuned & 32nd & 2/2 & 2/2 & 2/2 & 2/2 & -- \\
Web text, instruction-tuned & 128th & 2/2 & 2/2 & 2/2 & 2/2 & -- \\
\bottomrule
\end{tabular}
\end{table}

\section{Detailed comparison with prior studies}
\label{app:prior}

This appendix expands \cref{sec:related}, study by study, and summarizes the theory behind the training origin proposed in \cref{sec:discussion}.

\paragraph{The band at the final layers.} \citet{cancedda2024spectral} studied a band defined as ours but without vocabulary centring: the 5\% of the right singular directions of the unembedding, with the normalization weight folded in, that have the smallest singular values. Cancedda filtered the residual stream after one layer at a time in LLaMA-2 models with nested spectral filters and compared the loss with that of filters on random orthonormal bases of the same rank. From a swap experiment, Cancedda concluded that the band's primary role inside the network is to support attention sinks. In the final-layer rows of these loss maps (Cancedda's Figures~8 and~19), moreover, removing the band raises the loss more than removing a random subspace of the same rank. In this uncentred basis, our equal-length comparison reproduces the full result in every model (\cref{app:robust-equal}).

\paragraph{Normalization-mediated confidence.} \citet{stolfo2024confidence} defined the effective null space as the bottom singular directions of the unembedding with the LayerNorm weight folded in and the vocabulary mean removed, which is our unembedding. They showed that entropy neurons, which write into it, act partly through the LayerNorm scale: the scale carries about 80\% of their effect in GPT-2 and LLaMA-2 7B, and about 40\% in Pythia. Moreover, when the normalization is removed from GPT-2 models by fine-tuning, ablating these neurons no longer changes the loss \citep{baroni2026small}. Our change, by contrast, replaces the state's whole component in the band rather than these neurons' outputs, and it holds the scale fixed. It therefore measures the direct route for the whole component. On the natural path, where the scale is recomputed, the band's prediction change and loss change still exceed those of the references (\cref{app:robust-equal}).

\paragraph{The top token under HARP's removal.} \citet{hu2026harp} remove the unembedding's bottom right singular directions, 5\% of them by default and 256 in their best detection setting, and report that the top rankings of greedily generated tokens are maintained. \citet{joshi2025calibration} find task accuracy largely unchanged after removing the smallest 5 to 15\% of singular values.

We repeat the equal-length comparison in Llama-3.1-8B, which HARP lists as a backbone, in HARP's raw-unembedding basis (prespecified). This intervention removes the normalized state's component along the bottom 5\% of right singular directions, HARP's default. The removal's prediction change is 4.3 times that of equal-length random removals, and its loss excess is positive. Under HARP's metric, the unchanged top token stays first in 47\% of documents, against 68\% under the random removals (descriptive).

In the two Qwen base models, HARP's subspace carries on average 47\% (Qwen3-1.7B) and 69\% (Qwen3-8B) of the post-norm state's squared norm, against 24\% in Llama-3.1-8B. Its removal and the equal-length random removals are therefore large changes, and both rarely leave the unchanged top token first: in 15 to 16\% of documents after HARP's removal and in 5 to 13\% after the random removals (descriptive). By this share, HARP's removal looks the more benign. By the top token's median rank, however, it does not: 3{,}434 (Qwen3-1.7B) and 37{,}732 (Qwen3-8B) after HARP's removal, against 57 and 1{,}110 after the random removals. Its prediction change is nevertheless 2.0 (Qwen3-1.7B) and 1.3 (Qwen3-8B) times that of the random removals, and its loss excess is positive. Under our change, by contrast, the bottom band changes the most probable token more often than equal-length random changes in all ten models (\cref{app:band-consequences}).

\paragraph{Small singular values of internal weights.} In a 12-layer BERT fine-tuned on a sentence-similarity task, truncating the smallest 10\% of each layer's singular values reduced task performance as much as truncating the second-largest 10\% \citep{hsu2022fwsvd}. Similarly, removing the smallest tenth of the singular components of some weight matrices can raise perplexity or reduce task accuracy more than removing larger ones, depending on the matrix type and on fine-tuning \citep{staats2025small}. \citet{staats2025small} also find that, where the spectrum departs from random-matrix predictions at either end, the singular vectors overlap with the principal directions of the activations.

\paragraph{Matched updates during training.} A comparison at matched size also appears in \citet{godey2026lost}, who compare the loss change of two logit-update directions at the same logit-update norm. Their object is the gradient that flows back through the unembedding during training, whereas ours is a forward change of the final hidden state.

\paragraph{Predictive criteria for compression.} Beyond the Fisher-weighted criteria of \cref{sec:related}, low-rank compression also selects singular components by their predicted loss change in whitened coordinates \citep{abbasi2026zero}. Quantization, similarly, weights its objective by gradients of the end loss \citep{kim2025guidedquant} or matches the output distribution when it quantizes the final block \citep{lee2026lfq}. In compressed models, moreover, unchanged aggregate accuracy can conceal per-example changes that divergence measurements reveal \citep{dutta2024accuracy}. At the unembedding, finally, SVD-Softmax previews the logits with the leading singular components and recomputes the most probable words exactly, because the preview alone does not give accurate results \citep{shim2017svdsoftmax}. Our contrast shares, in turn, measure the part of the leading pair's contrast that lies in the bands such a preview omits.

\paragraph{Fidelity criteria for interpretability.} Besides the loss-based and divergence-based objectives of \cref{sec:related}, sparse autoencoder latents can be selected by the gradient of the output \citep{shu2025beyond}. Conclusions also depend on how fidelity is measured: CKA can miss the deletion of principal components beyond the leading ten, although probing accuracy drops \citep{ding2021grounding}.

\paragraph{The expressivity of the softmax output.} Work on the expressivity of the softmax output concerns the distributions the head can represent. It covers the softmax bottleneck \citep{chang2022softmax} and tokens whose probability is bounded by that of other tokens \citep{demeter2020stolen} or that can never be the argmax \citep{grivas2022low}. Our question, by contrast, concerns the consequence of a change of the state for a given prediction.

\paragraph{Implicit bias of next-token prediction.} In linear models with fixed context embeddings, gradient descent fits the logit differences among the tokens that can follow a context to their log-odds \citep{thrampoulidis2024implicit}. In the orthogonal subspace, the parameters grow in norm along a direction that separates those tokens from the rest. With unconstrained embeddings, similarly, the low-rank part of the logits, which depends only on which tokens can follow, dominates as training progresses \citep{zhao2024implicit}.

\end{document}